\pdfoutput=1
\documentclass[journal]{IEEEtran}
\usepackage{graphicx}
\usepackage{subfiles}
\usepackage{bm}
\usepackage{color}
\usepackage{soul}
\usepackage{amssymb}
\usepackage{amsmath}
\usepackage{url}
\usepackage{listings}

\title{Multivariate Uncertainty in Deep Learning}
\author{Rebecca L. Russell and Christopher Reale %
    \thanks{This work was carried out with funding from DARPA/MTO (HR0011-16-S-0001). Any opinions, findings and conclusions or recommendations expressed in this material are those of the authors.

    The authors are with The Charles Stark Draper Laboratory, Inc., Cambridge, MA 02139 (e-mail: rrussell@draper.com; creale@draper.com).}}

\begin{document}

\maketitle

\begin{abstract}
Deep learning has the potential to dramatically impact navigation and tracking state estimation problems critical to autonomous vehicles and robotics.
Measurement uncertainties in state estimation systems based on Kalman and other Bayes filters are typically assumed to be a fixed covariance matrix.
This assumption is risky, particularly for ``black box'' deep learning models, in which uncertainty can vary dramatically and unexpectedly.
Accurate quantification of multivariate uncertainty will allow for the full potential of deep learning to be used more safely and reliably in these applications.
We show how to model multivariate uncertainty for regression problems with neural networks, incorporating both aleatoric and epistemic sources of heteroscedastic uncertainty.
We train a deep uncertainty covariance matrix model in two ways: directly using a multivariate Gaussian density loss function, and indirectly using end-to-end training through a Kalman filter.
We experimentally show in a visual tracking problem the large impact that accurate multivariate uncertainty quantification can have on Kalman filter performance for both in-domain and out-of-domain evaluation data.
We additionally show in a challenging visual odometry problem how end-to-end filter training can allow uncertainty predictions to compensate for filter weaknesses.
\end{abstract}

\section{Introduction}
Despite making rapid breakthroughs in computer vision and other perception tasks, deep learning has had limited deployment in critical navigation and tracking systems. 
This lack of real-world usage is in large part due to challenges associated with integrating ``black box'' deep learning modules safely and effectively.
Navigation and tracking are important enabling technologies for autonomous vehicles and robotics, and have the potential to be dramatically improved by recent deep learning research in which physical measurements are directly regressed from raw sensor data, such as visual odometry \cite{mohanty2016deepvo}, object localization \cite{ren2015faster}, human pose estimation \cite{toshev2014deeppose}, object pose estimation \cite{wu2018real}, and camera pose estimation~\cite{kendall2015posenet}.

Proper uncertainty quantification is an important challenge for applications of deep learning within these systems, which typically rely on probabilistic filters, such as the Kalman filter~\cite{kalman1960}, to recursively estimate a probability distribution over the system's state from uncertain measurements and a model of the system evolution.
Accurate estimates of the uncertainty of a neural network ``measurement'' (i.e., prediction) would enable the integrated system to make better-informed decisions based on the fusion of measurements over time, measurements from other sensors, and prior knowledge of the underlying system.
The conventional approach of using a fixed estimate of measurement uncertainty can lead to catastrophic system failures when prediction errors and correlations are dynamic, as is often the case with deep learning perception modules.
By accurately quantifying uncertainty that can vary from sample to sample, termed \emph{heteroscedastic} uncertainty, we can create systems that gracefully handle deep learning errors while fully leveraging its strengths.

\begin{figure}[tb]
    \begin{center}
            \includegraphics[width=0.48\textwidth]{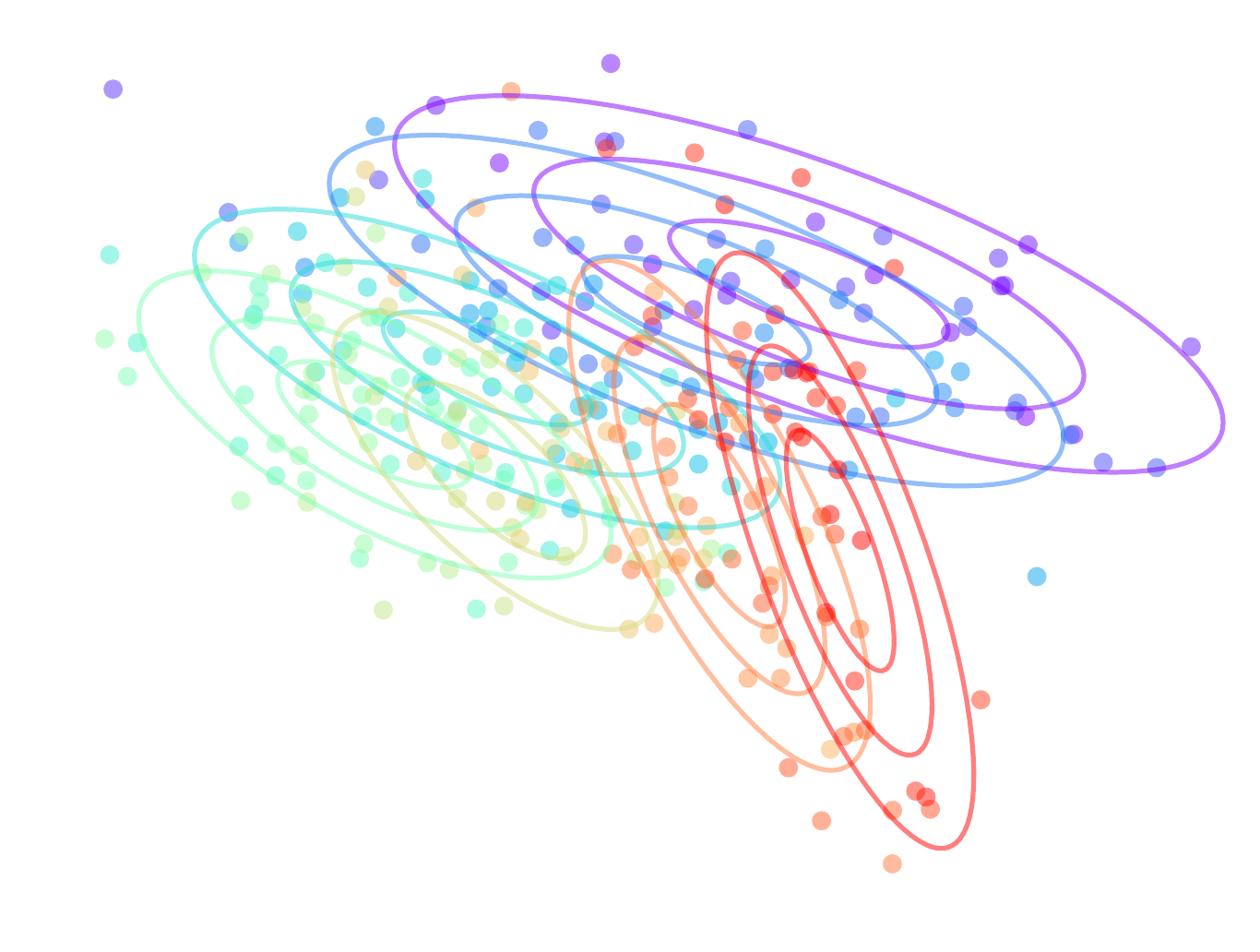}\vspace{-1cm}
    \end{center}
    \caption{Data from a vector function with heteroscedastic covariance}
    \label{fig:rainbow_contour}
\end{figure}

In this work, we study the quantification of heteroscedastic and correlated multivariate uncertainty (illustrated in Figure~\ref{fig:rainbow_contour}) for regression problems with the goal of improving overall performance and reliability of systems that rely on probabilistic filters.
Heteroscedastic uncertainty in deep learning can be modeled from two sources: \emph{epistemic} uncertainty and \emph{aleatoric} uncertainty~\cite{kendall2017}.
\emph{Epistemic} uncertainty reflects uncertainty in the model parameters and has been addressed by recent work to develop fast approximate Bayesian inference for deep learning~\cite{blundell2015weight, gal2016dropout, pearce2018uncertainty}.
Accurate estimation of epistemic uncertainty enables systems to perform more reliably in out-of-domain situations where deep learning performance can dramatically degrade.
\emph{Aleatoric} uncertainty reflects the noise inherent to the data and is irreducible with additional training.
Accurate estimation of aleatoric uncertainty enables systems to achieve maximum performance and most effectively fuse deep learning predictions.
Finally, since the uncertainties of predictions of multiple values can be highly \emph{correlated}, it is important to account for the full multivariate uncertainty from both aleatoric and epistemic sources.
Existing methods for uncertainty quantification in deep learning generally neglect correlations in uncertainty, though these correlations are often critically important in autonomy applications.

We show how to model multivariate aleatoric uncertainty through direct training with a special loss function in Section~\ref{sec:MLE} and through guided training via backpropagation through a Kalman filter in Section~\ref{sec:kalmantrain}.
In Section~\ref{sec:epistemic}, we show how to incorporate multivariate epistemic uncertainty, building off of the existing body of Bayesian deep learning literature. 
Finally, in Section~\ref{sec:kalman_exp}, we experiment with our techniques on a synthetic visual tracking problem that allows us to evaluate performance in-domain and out-of-domain test data and on a real-world visual odometry dataset with strong correlations between measurements.

\section{Related Work}
\label{sec:relwork}
Heteroscedastic noise is an important topic in the filtering literature. The adaptive Kalman filter~\cite{mehra1970identification} accounts for heteroscedastic noise by estimating the process and measurement noise covariance matrices online. This approach prevents the filter from keeping up with rapid changes in the noise profile.
In contrast, multiple model adaptive estimation \cite{magill1965optimal} uses a bank of filters with different noise properties and dynamically chooses between them, which can work well when there are a small number of regimes with different noise properties.
Covariance estimation techniques for specific applications, such as the iterative closest point algorithm \cite{censi2007accurate} and simultaneous localization and mapping \cite{pupilli2006real}, have been developed but do not generalize well.
 
Parametric~\cite{hu2015parametric} and non-parametric~\cite{kersting2007most, wilson2010generalised, vega2013cello, tallavajhula2016} machine learning methods, including neural networks~\cite{liu2018deep}, have been used to model aleatoric heteroscedastic noise from sensor data.
Most of these approaches scale poorly and, since they ignore epistemic uncertainty, cannot be used with measurements inferred through deep learning.
Kendall and Gal \cite{kendall2017} showed how to predict the variance of neural network outputs including epistemic uncertainty, but neglected correlations between the uncertainty of different outputs and how these correlations might affect downstream system performance.

Several recent works have also investigated the direct learning of neural network models~\cite{haarnoja2016backprop, jonschkowski2018differentiable}, including measurement variance models~\cite{coskun2017long}, via backpropagation through Bayes filters.
These works demonstrated the practicality and power of filter-based training, but none attempted to account for the epistemic uncertainty or full multivariate aleatoric uncertainty of the neural networks.
Additionally, the impact of improved uncertainty quantification, rather than simply improved measurement and process modeling, has not yet been studied for deep learning in probabilistic filters.

We build upon this prior work by showing how to predict multivariate uncertainty from both epistemic and aleatoric sources without neglecting correlations. Furthermore, we show how to do this by either training through a Kalman filter or independently from one.
Our work is the first to show how to comprehensively quantify deep learning uncertainty in the context of Bayes filtering systems that are dependent on the outputs of deep learning models.

\section{Multivariate Uncertainty Prediction}
\label{sec:theory}
\begin{table}[]
\begin{center}
\caption{\label{tab:notation}Notation}
\begin{tabular}{|c|l|}
\hline
 $\bm{x} \in \mathcal{X}$ & Neural network model input \\ \hline
 $\bm{y} \in \mathbb{R}^k$ & Regression label for neural network \\ \hline
    $\bm{f}: \mathcal{X} \to \mathbb{R}^k$ & Neural network model of expected $\bm{y}$ mean \\ \hline
    $\bm{\Sigma}: \mathcal{X} \to \mathbb{R}^{k\times k}$ & Neural network model of expected $\bm{y}$ covariance \\ \hline
    $\bm{z} \in \mathbb{R}^n $ & Filter system state \\ \hline
    $\bm{\hat{z}} \in \mathbb{R}^n $ & Estimate of system state $\bm{z}$ \\ \hline
    $\mathbf{P} \in \mathbb{R}^{n\times n} $ & Estimate of system state $\bm{z}$ covariance \\ \hline
    $\mathbf{F}: \mathbb{R}^n \to \mathbb{R}^n$ & State-transition model (fixed, linear) \\ \hline
    $\mathbf{H}: \mathbb{R}^n \to \mathbb{R}^k$ & Observation model (fixed, linear) \\ \hline
    $\mathbf{Q} \in \mathbb{R}^{n\times n} $ & Process noise covariance (fixed) \\ \hline
\end{tabular}
\end{center}
\end{table}

We present two methods for training a neural network to predict the multivariate uncertainty of either its own regressed outputs or those of another measurement system.
The first is based on direct training using a Gaussian maximum likelihood loss function (Section \ref{sec:MLE}) and the second is indirect end-to-end training through a Kalman filter (Section \ref{sec:kalmantrain}).
These two methods can be either used alone or in conjunction (using direct training as a pre-training step before end-to-end training), depending on the exact application and availability of labeled data.
For training a neural network to estimate its own uncertainty, we also present a method to approximately incorporate epistemic uncertainty at test time (Section \ref{sec:epistemic}).
Table~\ref{tab:notation} summarizes the important notation used in this section and throughout the rest of the paper.
\subsection{Gaussian maximum likelihood training} \label{sec:MLE}

In this first method, we directly learn to predict covariance matrix parameters that describe the distribution of training data labels with respect to the corresponding predictions of them.

We assume that the probability of a label $\bm{y} \in \mathbb{R}^k $ given a model input $\bm{x} \in \mathcal{X}$ can be approximated by a multivariate Gaussian distribution
\begin{equation}
    \begin{aligned}
        p\left(\bm{y} \mid \bm{x}\right) =& \frac{1}{\sqrt{(2\pi)^k \left|\bm{\Sigma}(\bm{x})\right|}}\times {}\\
        & \exp\left[-\frac{1}{2} \left(\bm{y}-\bm{f}(\bm{x})\right)^T \bm{\Sigma}({\bm{x}})^{-1} \left(\bm{y}-\bm{f}(\bm{x})\right)\right],
    \end{aligned}
    \label{eq:gauss}
\end{equation}
where $\bm{f}: \mathcal{X} \to \mathbb{R}^k$ is a model of the mean, $\mathbb{E}\left[\bm{y}\,|\,\bm{x}\right]$, and $\bm{\Sigma}: \mathcal{X} \to \mathbb{R}^{k\times k}$ is a model of the covariance, 
\begin{equation}
    \mathbb{E}\left[\left.\left(\bm{y}-\bm{f}(\bm{x}\right)\left(\bm{y}-\bm{f}(\bm{x})\right)^T\right|\bm{x}\right].
    \label{eq:covexp}
\end{equation}
To train $\bm{f}$ and $\bm{\Sigma}$, we find the parameters that minimize our loss $\mathcal{L}$, the negative logarithm of the Eq.~\ref{eq:gauss} likelihood,
\begin{equation}
\mathcal{L} = \frac{1}{2}\left(\bm{y}-\bm{f}(\bm{x})\right)^T\bm{\Sigma}(\bm{x})^{-1}\left(\bm{y}-\bm{f}(\bm{x})\right) + \frac{1}{2}\ln\left|\bm{\Sigma}(\bm{x})\right|.
\label{eq:loss_func}
\end{equation}
This loss function allows us to train $\bm{f}$ and $\bm{\Sigma}$ either simultaneously or separately.
Typically, we use a single base neural network that outputs $\bm{f}$ and $\bm{\Sigma}$ simultaneously.

The $\bm{\Sigma}$ model should output $k$ values $\bm{s}$, which we use to define the variances along the diagonal
    \begin{equation}
        \Sigma_{ii} = \sigma_i^2 = g_v(s_i)
    \end{equation}
and $k(k-1)/2$ additional values $\bm{r}$, which, along with $\bm{s}$, define the off-diagonal covariances
\begin{equation}
    \Sigma_{ij} = \rho_{ij} \sigma_i \sigma_j = g_\rho(r_{ij})\sqrt{g_v(s_i)g_v(s_j)},
\end{equation}
where $\Sigma_{ij}=\Sigma_{ji}$ for $j<i$. We use the $g_v = \exp$ activation for the variances, $\sigma^2_i$, and $g_\rho = \tanh$ activation for the Pearson correlation coefficients, $\rho_{ij}$, to stabilize training and help encourage prediction of valid positive-definite covariance matrices.

\subsection{Kalman-filter training}\label{sec:kalmantrain}

\begin{figure}
    \begin{center}
        \includegraphics[width=0.48\textwidth]{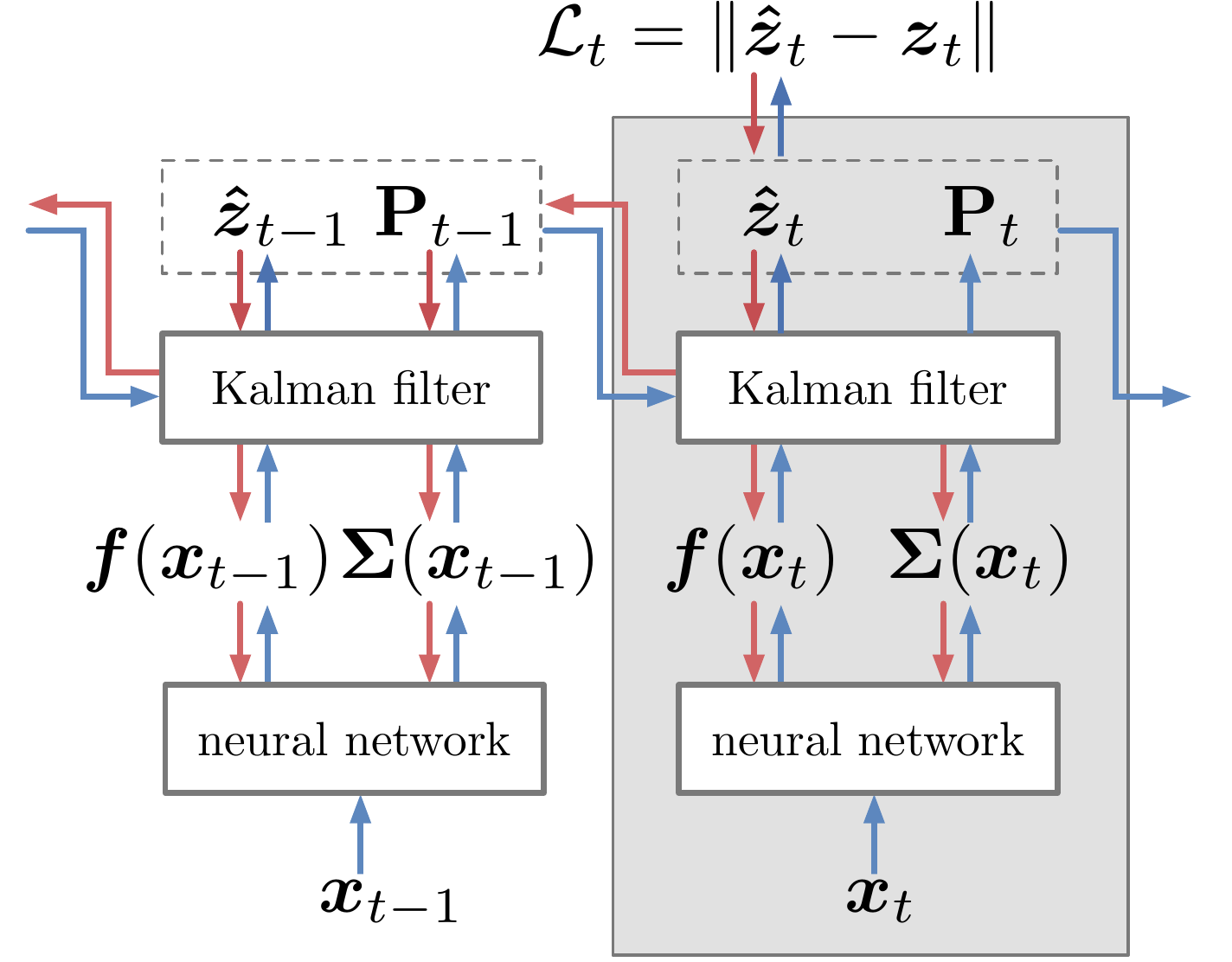}
    \end{center}
    \caption{Simultaneously training a neural network via a Kalman filter to output a measurement $\bm{f}$ and its measurement error covariance $\bm{\Sigma}$ based on some input $\bm{x}$. At each time step $t$, the Kalman filter calculates a state estimate $\bm{\hat{z}}_t$ and error covariance matrix $\textbf{P}_t$ based on the previous state estimate and its covariance and the new measurement and its covariance. The loss contribution $\mathcal{L}_t$ at time step $t$ from the Kalman filter state estimate thus depends on the current and all previous outputs of $\bm{f}$ and $\bm{\Sigma}$. Blue arrows show the forward propagation of information and red arrows show the backpropagation of gradients that train the neural network.}
    \label{fig:kalman_train}
\end{figure}

Our second method of training a neural network to predict multivariate uncertainty uses indirect training through a Kalman filter, illustrated in Figure~\ref{fig:kalman_train}.
The Kalman filter~\cite{kalman1960} is a state estimator for linear Gaussian systems that recursively fuses information from measurements and predicted states.
The relative contribution of these two sources is determined by their covariance estimates.
Similarly to Section \ref{sec:MLE}, we use a neural network to predict the heteroscedastic measurement covariance, but instead of training with the Eq.~\ref{eq:loss_func} loss function, we train using the error of the Kalman state estimate.

The standard Kalman filter assumes that we can specify a state-transition model $\mathbf{F}$ and observation model $\mathbf{H}$.
The state-transition model describes the evolution of the system state $\bm{z}$ by
\begin{equation}
    \bm{z}_t = \mathbf{F}\bm{z}_{t-1} + \bm{w}_t, 
\end{equation}
where $\bm{w}_t \sim\mathcal{N}(\bm{0}, \mathbf{Q})$ and $\mathbf{Q}$ is the covariance of the process noise.
We model the relationship between our deep learning ``measurement'' $\bm{f}(\bm{x}_t)$ and the system state as
\begin{equation}
    \bm{f}(\bm{x}_t) = \mathbf{H}\bm{z}_{t} + \bm{v}_t, 
\end{equation}
where $\bm{v}_t \sim\mathcal{N}(\bm{0}, \mathbf{\Sigma}(\bm{x}_t))$ and $\mathbf{\Sigma}(\bm{x}_t)$ is the covariance of the measurement noise.
The difference between the measurement and the predicted measurement is the \emph{innovation}
\begin{equation}
    \bm{i}_t = \bm{f}(\bm{x}_t) - \mathbf{H}\,\mathbf{F} \hat{\bm{z}}_{t-1},
\end{equation}
which has covariance
\begin{equation}
    \mathbf{S}_t = \mathbf{\Sigma}(\bm{x}_t) + \mathbf{H}\left(\mathbf{F}\,\mathbf{P}_{t-1}\mathbf{F}^T + \mathbf{Q}\right)\mathbf{H}^T,
    \label{eq:icov}
\end{equation}
the sum of the measurement covariance and the predicted covariance from the previous state estimate, $\mathbf{P}_{t-1}$.

We can then update our state estimate based on the innovation by
\begin{equation}
    \hat{\bm{z}}_t = \mathbf{F}\hat{\bm{z}}_{t-1} + \mathbf{K}_t \bm{i}_t,
\end{equation}
where $\mathbf{K}_t$ is the \emph{Kalman gain}, which determines how much the state estimate should reflect the current measurement compared to previous measurements.
The optimal Kalman gain that maximizes the likelihood of the true state $\bm{z}_t$ given our Gaussian assumptions is
\begin{equation}
    \mathbf{K}_t = \mathbf{F}\,\mathbf{P}_{t-1}(\mathbf{H}\, \mathbf{F})^T \mathbf{S}_t^{-1}.
    \label{eq:kgain}
\end{equation}
The covariance $\mathbf{P}_t$ of the updated state estimate $\hat{\bm{z}}_t$ is then given by
\begin{equation}
    \mathbf{P}_t = \left(\mathbf{I} - \mathbf{K}_t \mathbf{H}\right)\left(\mathbf{F}\mathbf{P}_{t-1}\mathbf{F}^T + \mathbf{Q}\right).
\end{equation}

Our measurement covariance $\bm{\Sigma}$ enters the Kalman filter in the calculation of the innovation covariance $\mathbf{S}_t$ in Eq.~\ref{eq:icov} and thus directly but inversely affects the Kalman gain $\mathbf{K}_t$ in Eq.~\ref{eq:kgain}.
We can therefore backpropagate errors from the state estimate ${\hat{\bm{z}}}_t$ through the Kalman filter to train the model for $\bm{\Sigma}$.
In fact, we can use any subset $S$ of the state indices to train the measurement covariance model so long as
\begin{equation}
    \sum_{b \in S} H_{ab} \neq 0\qquad \forall\, a \in \{1, ..., k\},
\end{equation}
which is useful when full state labels are not available, as is often the case in more complex systems.
Likewise, we can easily learn to predict the covariance of just a subset of the measurement space, for example, when the filter is fusing other well-characterized sensors with a neural network prediction.
We rely on automatic differentiation to handle the calculation of the fairly complicated gradients through the filter.

If all the assumptions for the Kalman filter hold, $\bm{\Sigma}$ will converge to Eq.~\ref{eq:covexp} maximizing the likelihood of $\{\bm{z}_t\}$, theoretically equivalent to the method in Section~\ref{sec:MLE}.
In that case, the main benefit to using Kalman filter training is to allow for labels other than $\{\bm{y}\}$ to be used in training.
In situations where the Kalman filter assumptions are violated and the Kalman filter is the desired end-usage, it can be preferable to optimize for the overall end-to-end system performance.
As demonstrated experimentally in Section~\ref{sec:kalman_exp}, this end-to-end training allows the model of $\bm{\Sigma}$ to compensate for other modeling weaknesses.
On the other hand, training through a Kalman filter can be slow or prone to instability for many applications, so the more direct approach Section~\ref{sec:MLE} is preferable when the appropriate labels are available, at the minimum as a pre-training stage.

The approach shown here for the standard Kalman filter can be extended to other Bayes filters including non-linear and sampling-based variants, allowing use in nearly any state estimation application.

\subsection{Incorporation of epistemic uncertainty}\label{sec:epistemic}

Our two uncertainty prediction methods developed in Sections \ref{sec:MLE} and \ref{sec:kalmantrain} quantify aleatoric uncertainty in the training data independently of epistemic uncertainty.
Epistemic uncertainty, also known as ``model uncertainty'', represents uncertainty in the neural network model parameters themselves.
Epistemic uncertainty for neural networks models is reduced through the training process, allowing us to isolate the contributions of aleatoric uncertainty in the training data in both of our aleatoric uncertainty prediction methods.
Like aleatoric uncertainty, epistemic uncertainty can vary dramatically from measurement-to-measurement.
Epistemic uncertainty is a particular concern for neural networks given their many free parameters, and can be large for data that is significantly different from the training set.
Thus, for any real-world application of neural network uncertainty estimation, it is critical that it be taken into account.

Numerous sampling-based approaches for Bayesian inference have been developed that allow for the estimation of epistemic uncertainty, all of which are compatible with our uncertainty quantification framework.
The easiest and most practical approach is to use dropout Monte Carlo~\cite{gal2016dropout}, which trades off accuracy for speed and convenience.
Recent Bayesian ensembling approaches~\cite{pearce2018uncertainty} driven by the empirical success of ensembling for estimating uncertainty~\cite{lakshminarayanan2017} are also promising.

As in Ref.~\cite{kendall2017}, we assume that aleatoric and epistemic uncertainty are independent.
We train a neural network to output the aleatoric uncertainty covariance $\bm{\Sigma}(\bm{x})$ of the prediction $\bm{f}(\bm{x})$ through either method presented in Sections \ref{sec:MLE} and \ref{sec:kalmantrain}.
Then, using any sampling-based method for the estimation of epistemic uncertainty from $N$ samples of $\bm{f}(\bm{x})$, the total predictive covariance estimate should be calculated by
\begin{equation}
    \label{eq:sigma_pred}
    \begin{split}
        \bm{\Sigma}_\textit{pred} =& \bm{\Sigma}_\textit{epistemic} + \bm{\Sigma}_\textit{aleatoric} \\
        \approx& \frac{1}{N} \sum_{n=1}^N f_n(\bm{x})f_n(\bm{x})^T \\
        &-\frac{1}{N^2}\left(\sum_{n=1}^N f_n(\bm{x})\right)\left(\sum_{n=1}^N f_n(\bm{x})\right)^T \\
        &+ \frac{1}{N}\sum_{n=1}^N \bm{\Sigma}_n (\bm{x}).
    \end{split}
\end{equation}

As long as epistemic uncertainty for the training data is small relative to aleatoric uncertainty, this formulation only needs to be used at test time.
However, if epistemic uncertainty is significant for data in the training set after training, the predicted $\bm{\Sigma}$ directly from the neural network will incorporate this uncertainty in its prediction, making it challenging to separate.
We found it possible to calculate the epistemic uncertainty covariances for the training data set and \emph{tune} the model covariance prediction to predict the residual from that with Eq.~\ref{eq:loss_func} while holding the main model $\bm{f}$ constant.
However, this tuning is less stable and our best results were generally achieved by training until the epistemic uncertainty had become small relative to aleatoric uncertainty for the training set.

\section{Experiments}
\label{sec:kalman_exp}
\begin{figure*}
    \begin{center}
        \includegraphics[width=0.98\textwidth]{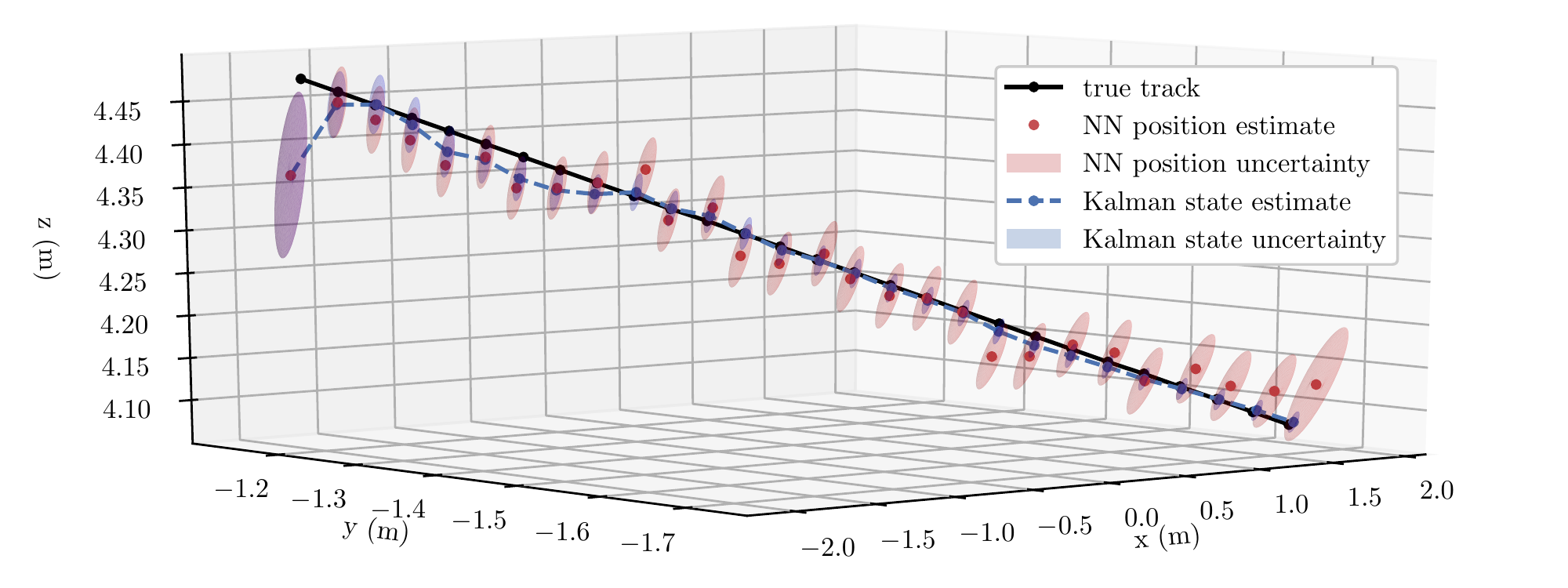}
    \vspace{-.2cm}
    \end{center}
    \caption{Kalman filter state estimates (blue) from the neural network's position and position covariance predictions (red) for the 3D visual tracking problem. As additional observations are made (left to right), the Kalman filter state estimate approaches the true track (black).}
    \label{fig:example_track}
\end{figure*}

We evaluate our methods developed in Section~\ref{sec:theory} on two practical use cases. First, we run experiments on a synthetic 3D visual tracking dataset that allows us to investigate the impact of aleatoric and epistemic uncertainty for in-domain and out-of-domain data.
Second, we study our methods on a real-world visual odometry dataset with strong correlations between measurements and complex underlying dynamics.

To provide a fair comparison between different uncertainty prediction methods, for each experiment, we freeze the prediction results $\bm{f}(\bm{x})$ from a trained model and then tune the individual uncertainty quantification methods on these prediction results from the training dataset.
We compare four methods for representing the multivariate uncertainty $\bm{\Sigma}$ from a neural network in a Kalman filter:
\begin{enumerate}
    \item \emph{Fixed covariance}: As a baseline, we calculate the measurement error covariance over the full dataset. This is the conventional approach to estimating $\bm{\Sigma}$ for a Kalman filter.
    \item \emph{MLE-learned variance}: The covariance prediction is learned assuming assuming no correlation between outputs, equivalent to the loss function given in Ref.~\cite{kendall2017}.
    \item \emph{MLE-learned covariance}: The covariance prediction is learned using the Eq.~\ref{eq:loss_func} loss function, as described in Section~\ref{sec:MLE}.
    \item \emph{Kalman-learned covariance}: The covariance prediction is learned through training via Kalman filter, as described in Section~\ref{sec:kalmantrain}.
\end{enumerate}
Finally, we compare these results to using a fully-supervised recurrent neural network (RNN) instead of a Kalman filter, to provide a point of reference for an end-to-end deep learning approach that does not explicitly incorporate uncertainty or prior knowledge.

\subsection{3D visual tracking problem}\label{sec:tracking}
We first consider a task that is simple to simulate but is a reasonable proxy for many practical, complex applications: 3D object tracking from video data.
For this problem, we trained a neural network to regress the $x$-$y$-$z$ position of a predetermined object in a single image, and used a Kalman filter to fuse the individual measurements over the video into a full track with estimated velocity.
This application allows the neural network to handle the challenging computer vision component of the problem, while the Kalman filter builds in our knowledge of physics, geometry, and statistics.
By using simulated data, we are able to carefully evaluate our methods of learning multivariate uncertainty on both in-domain and out-of-domain data.

We generated data using Blender \cite{blender} to render frames of objects moving through 3D space.
Each track was generated with random starting location, direction, and velocity (from 10 mm/s to 200 mm/s).
The tracks had constant velocity, though a more complex kinematic model or process noise could easily be added without changing the experiments significantly.
The object orientation was sampled uniformly independently for each frame in order to add an additional source of visual noise.
Our experiments used a ResNet-18~\cite{he2016} with the final linear layer replaced by a $512\times 512$ linear layer with dropout and size-3 position and size-6 covariance outputs.

An example of a Kalman filter being used on our visual tracking problem is shown in Figure~\ref{fig:example_track}, with both the neural network measurement uncertainty $\bm{\Sigma}$ and the Kalman state estimate uncertainty $\mathbf{P}$ at each frame shown.
We represent the state for our tracking problem as the 3D position and velocity of the object, using a constant-velocity state-transition model and no process noise.
During evaluation, the full $\bm{\Sigma}_\textit{pred}$ given in Eq. \ref{eq:sigma_pred} should be used to incorporate epistemic uncertainty in the Kalman filter's error handling.

\begin{figure*}
    \begin{center}
        \includegraphics[width=0.98\textwidth]{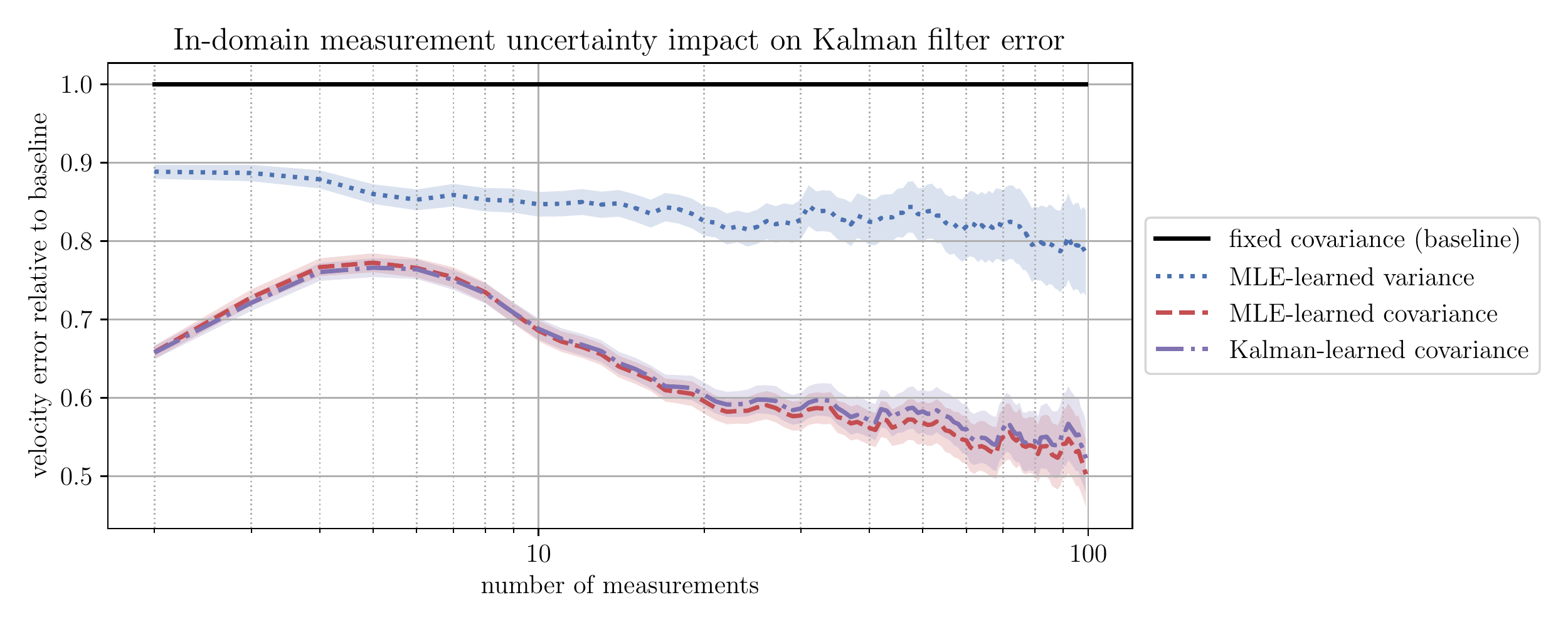}
    \end{center}
    \vspace{-.2cm}
    \caption{Improvement in Kalman filter velocity estimation in the 3D visual tracking problem as a function of measurement count for three uncertainty estimation methods. Large improvement over the conventional fixed measurement covariance approach is seen from accounting for both heteroscedasticity and correlation.}
    \label{fig:kalman_results}
\end{figure*}

Our four comparison methods of uncertainty quantification were evaluated using the track velocity estimation of the Kalman filter on a test set of \emph{in-domain} track data.
The results, shown in Table~\ref{tab:results}, indicate that moving from the fixed covariance to heteroscedastic covariance estimation yields a large improvement in the quality of the filter estimates.
Both learned covariance methods further dramatically improve the results, indicating that accounting for the correlations within the measurements can be very important.
The MLE-based and Kalman-based covariance learning methods were generally consistent with each other.
The improvement over the baseline fixed covariance method is plotted versus the number of tracked measurements in Figure~\ref{fig:kalman_results}.

\begin{figure*}
    \begin{center}
        \includegraphics[width=0.98\textwidth]{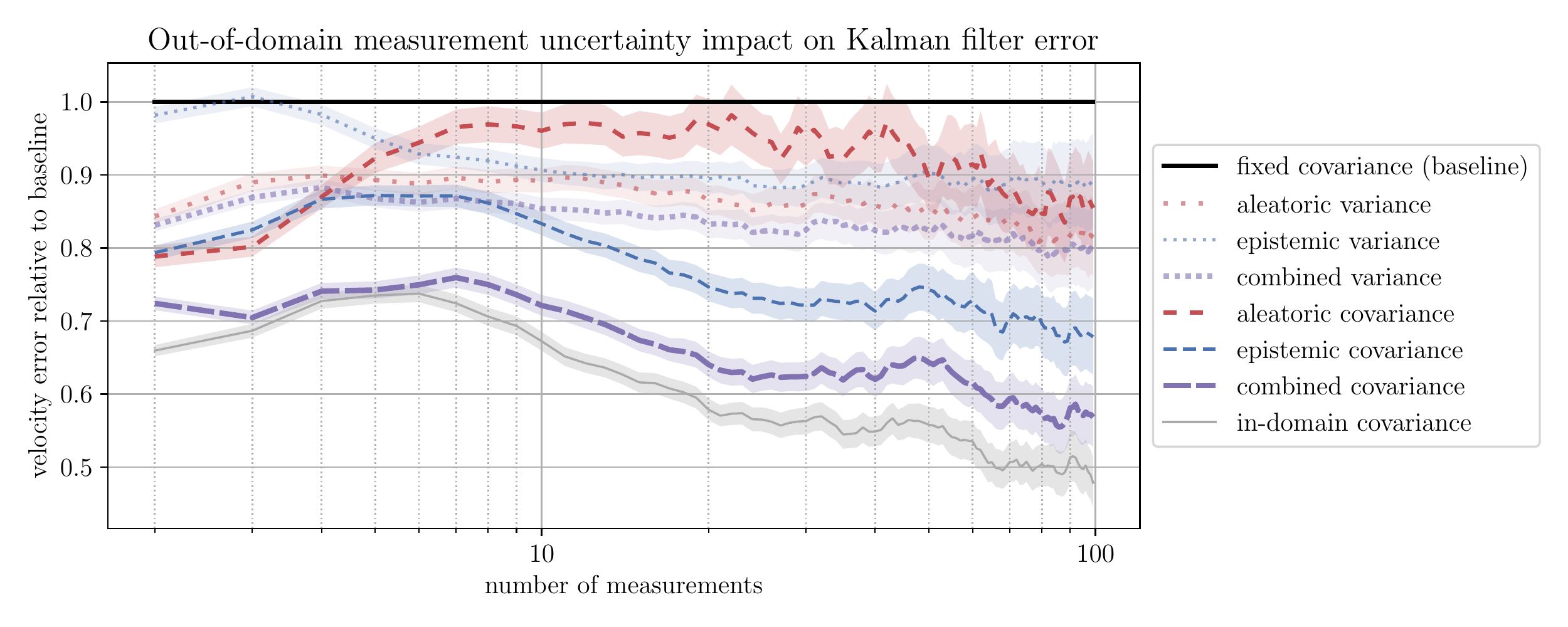}
    \end{center}
    \vspace{-.2cm}
    \caption{Improvement in Kalman filter velocity estimation in the 3D visual tracking problem as a function of measurement count for three uncertainty estimation methods. Large improvement over the conventional fixed measurement covariance approach is seen from accounting for both heteroscedasticity and correlation.}
    \label{fig:kalman_results_ood}
\end{figure*}

To test the quality of the uncertainty estimation methods when epistemic uncertainty is significant, we simulated \emph{out-of-domain} data by randomly jittering the input image color channel, a form of data augmentation not seen during training.
The results for the MLE-trained variance and covariance, as well as their break down into aleatoric and epistemic uncertainty, are shown in Table~\ref{tab:results_ood} and Figure~\ref{fig:kalman_results_ood}.
The ``in-domain covariance'' results when just the uncertainty estimation model is trained on the out-of-domain position predictions are added to provide a best-case-scenario point of comparison.
When evaluated on out-of-domain data, the performances of the aleatoric-only uncertainty estimates are greatly diminished, and the incorporation of correlation into the estimation no longer seems to help.
However, when the epistemic and aleatoric uncertainties are \emph{combined}, the results are close to the in-domain best-case-scenario and accounting for the correlation in uncertainty again gives a large improvement.
These results illustrate how critical the incorporation of epistemic uncertainty is in real applications of neural networks, where data is not guaranteed to remain in-domain.

\begin{table}[tb]
\centering
\caption{\label{tab:results}In-domain tracking velocity estimation}
\begin{tabular}{ c | c c | c c}
    & \multicolumn{2}{c|}{error (mm/s)} & \multicolumn{2}{c}{relative error}\\
    uncertainty method & mean & median & mean & median \\ \hline
    fixed covariance (baseline) & 2.00 & 0.75 & 1 & 1 \\
    MLE-learned variance & 1.87 & 0.61 & 1.00 & 0.88 \\
    \textbf{MLE-learned covariance} & \textbf{1.36} & \textbf{0.32} & \textbf{0.70} & \textbf{0.51} \\
    \textbf{Kalman-learned covariance} & \textbf{1.37} & \textbf{0.32} & \textbf{0.72} & \textbf{0.53}
\end{tabular}
\vspace{0.5cm}
\caption{\label{tab:results_ood}Out-of-domain tracking velocity estimation}
\begin{tabular}{ c | c c | c c}
    & \multicolumn{2}{c|}{error (mm/s)} & \multicolumn{2}{c}{relative error}\\
    uncertainty method & mean & median & mean & median \\ \hline
    fixed covariance (baseline) & 2.14 & 0.74 & 1 & 1 \\
    aleatoric variance & 1.98 & 0.62 & 1.01 & 0.89 \\
    epistemic variance & 1.94 & 0.61 & 0.93 & 0.89 \\
    combined variance & 2.00 & 0.61 & 0.97 & 0.89 \\
    aleatoric covariance & 1.87 & 0.48 & 1.10 & 0.73 \\
    epistemic covariance & 1.65 & 0.43 & 0.76 & 0.67 \\
    \textbf{combined covariance} & \textbf{1.56} & \textbf{0.37} & \textbf{0.75} & \textbf{0.60} \\ \hline
    in-domain covariance & 1.46 & 0.32 & 0.71 & 0.53
\end{tabular}
\end{table}

To provide a final comparison, we replaced the Kalman filter with an RNN that was trained directly on the track velocity labels using the same pre-trained convolutional filters as before.
On the in-domain tracking velocity estimation task, after careful tuning, the RNN was able to achieve a mean error of 1.37 mm/s, equivalent to our multivariate uncertainty approaches.
This result indicates that the RNN was able to successfully learn the filtering process and implicitly incorporate aleatoric uncertainty in individual measurements.
However, on the out-of-domain velocity estimation task, the RNN achieved only a mean error of 2.05 mm/s, as it has no way to incorporate epistemic uncertainty in its estimates and is vulnerable to catastrophic prediction failures when input data is far out of domain.

\subsection{Real-world visual odometry problem}\label{sec:odometry}

In order to evaluate our methods on real-world visual imagery in another important practical application, we study monocular visual odometry using the KITTI Vision Benchmark Suite dataset~\cite{kitti}.
Monocular visual odometry is the estimation of camera motion from a single sequence of images, and is particularly challenging since stereo imagery or depth is typically needed to estimate the distance of features in the scene from the camera in a pair of images.
While the KITTI dataset provides both stereo images and laser ranging data, deep learning has the potential to perform effective visual odometry with a much simpler and cheaper set of sensors, enabling better localization for inexpensive autonomous vehicles and robots.

\begin{figure}
    \begin{center}
        \includegraphics[width=0.35\textwidth]{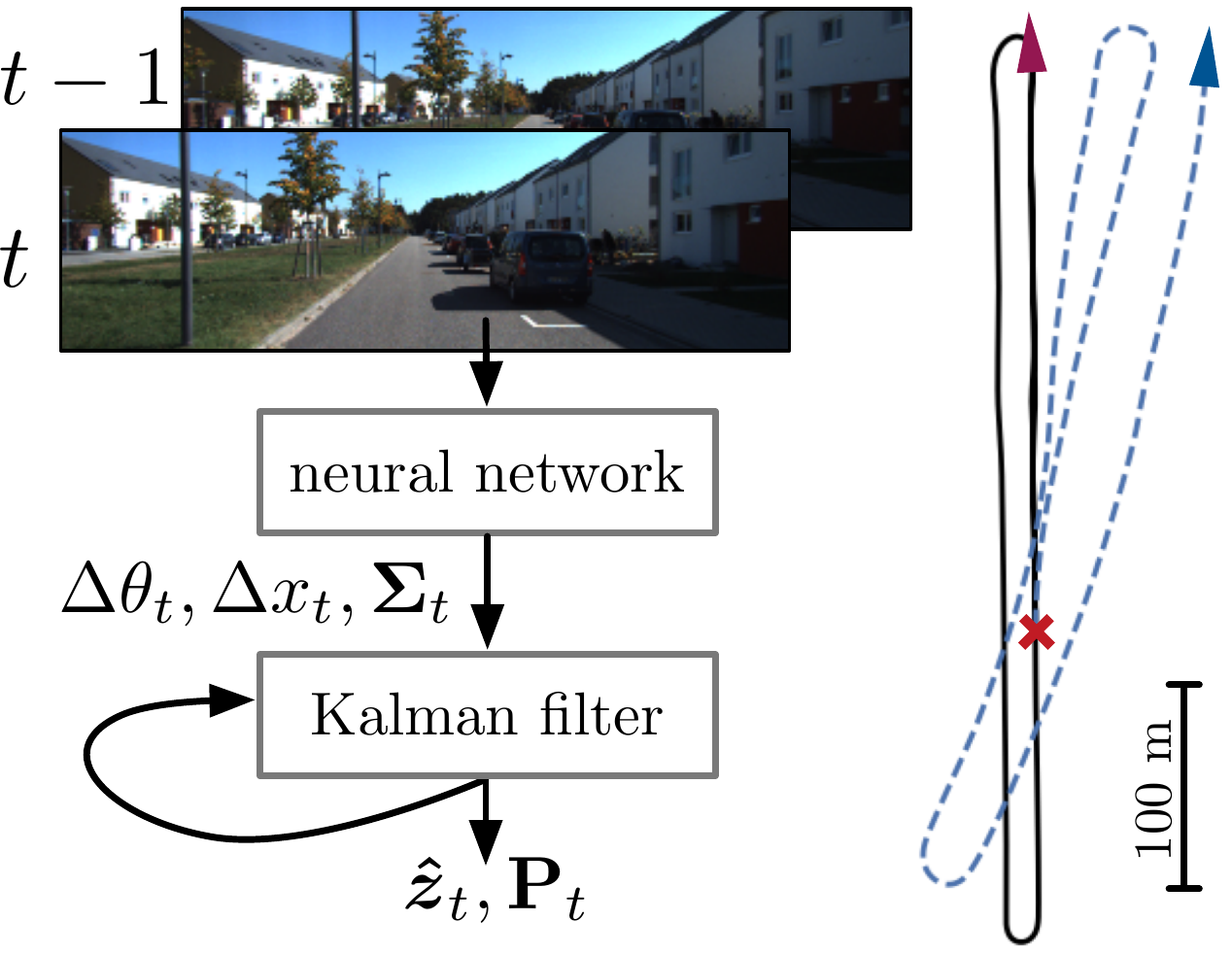}
    \vspace{-.2cm}
    \end{center}
    \caption{An illustration of the KITTI visual odometry problem, with data from validation run 06. Two adjacent video frames are input to the neural network, which predicts the change in angle ($\Delta \theta_t$) and position ($\Delta x_t$) between them. A Kalman filter fuses these predictions with previous predictions and other sensor measurements, resulting in a estimate (blue dashed line) of the true trajectory (black line).}
    \label{fig:kitti}
\end{figure}

We focus on the problem of estimating camera motion from a single pair of images using a neural network, as shown in Figure~\ref{fig:kitti}.
The neural network directly predicts the change in orientation ($\Delta\theta$) and position ($\Delta x$) of the camera, as well as its multivariate uncertainty $\bm{\Sigma}$.
The KITTI dataset contains driving data from a standard passenger vehicle in relatively flat areas, so we focus on predicting the vehicle speed and change in heading.
We filter these measurements with noisy IMU (inertial measurement unit) data, which reflect what would be available on an inexpensive platform, as well as with the previous measurements from both sensor sources.
The filtered estimate of $\Delta \theta$ can then be integrated to find the heading of the vehicle, combined with $\Delta x$ calculate the velocity components of the vehicle, and those components integrated to find the position of the vehicle, also illustrated in Figure~\ref{fig:kitti}.

For our neural network architecture and training, we follow the general approach of Wang \emph{et al.}~\cite{deepvo}, using a pre-trained FlowNet~\cite{flownet} convolutional neural network to extract features from input image pairs.
Instead of using a recurrent neural network to fuse these features over many images, we have two fully-connected layers that output $\Delta\theta_t$, $\Delta x_t$, and $\bm{\Sigma}_t$.
We train on KITTI sequences 00, 01, 02, 05, 08, and 09, and evaluate on sequences 04, 06, 07, and 10.

We evaluate results using two different filters: a standard Kalman filter and a Kalman filter that takes into account time-correlated measurement noise.
Both filters use a simple constant velocity and rotation model of the dynamics, which thus has a high corresponding process noise $\mathbf{Q}$ determined from the data.
The standard Kalman filter assumes that all measurements have \emph{independent} errors, a poor assumption for the neural network outputs, which are highly correlated between adjacent times.
The ``time-correlated'' Kalman filter~\cite{wang2012practical} accounts for this correlation by modeling the measurement error as part of the estimated state $\bm{\hat{z}}_t$ and combining the uncorrelated part of $\bm{\Sigma}_t$ into $\mathbf{Q}$.

\begin{table}[tb]
\centering
\caption{\label{tab:results_odom}Visual odometry experiment results}
\begin{tabular}{ c | c c | c c}
    & \multicolumn{2}{c|}{standard Kalman} & \multicolumn{2}{c}{time-correlated}\\
    uncertainty method & $\Delta\theta_{\mathrm{err}}$ & $\Delta x_{\mathrm{err}}$ & $\Delta\theta_{\mathrm{err}}$ & $\Delta x_{\mathrm{err}}$ \\ \hline
    fixed covariance (baseline) & 0.239 & 0.147 & 0.233 & 0.136 \\
    MLE-learned variance & 0.205 & 0.163 & 0.188 & 0.128 \\
    MLE-learned covariance & 0.183 & 0.162 & 0.163 & 0.129 \\
    \textbf{Kalman-learned covariance} & \textbf{0.171} & \textbf{0.138} & \textbf{0.155} & \textbf{0.128} \\
\end{tabular}
\end{table}

The results from our visual odometry experiments are shown in Table~\ref{tab:results_odom}, where $\Delta \theta_\mathrm{err}$ is the frame-pair angle error in degrees and $\Delta x_\mathrm{err}$ is the frame-pair position error in meters.
For the standard Kalman filter, which neglects the large measurement time-correlation, we find that the MLE-learned methods perform poorly but the Kalman-learned covariance is able to compensate by strategically increasing the predicted uncertainty for highly time-correlated measurements.
With the time-correlated Kalman filter variant, all uncertainty-quantification methods achieve better performance, and the MLE-learned covariance results are more in-line with the Kalman-learned covariance results.
Again, the Kalman-learned covariance is able to achieve the best performance out of all methods by directly optimizing for the desired end state estimate.

Finally, we provide a comparison to an end-to-end deep learning RNN method, equivalent to the approach of Wang~\emph{et al.}~\cite{deepvo} with the addition of an IMU input data.
The RNN is trained to fuse sequences of FlowNet features into odometry outputs, allowing it to incorporate learned knowledge about trajectory dynamics in addition to the fusion of correlated measurements in its ``filtering'' process.
The RNN was able to achieve a $\Delta \theta_\mathrm{err}$ of 0.155 degrees, equal to our Kalman-learned covariance approach, but a $\Delta x_\mathrm{err}$ of 0.176 meters, worse than any of our Kalman filter results.
We believe that the RNN was able to learn a very good trajectory model for vehicle heading (drive straight with occasional large turns).
The Kalman filter assumption of constant rotation was not a very good one for the data, explaining why the Kalman-learned covariance was able to improve upon the MLE-learned covariance so much for $\Delta \theta_\mathrm{err}$.
On the other hand, the constant velocity assumption was a relatively good one and robust for trajectories with high epistemic uncertainty, such as when test trajectories are significantly faster or slower than typical training data.
This result illustrates a primary strength of our uncertainty quantification approach: It allows traditional filtering methods to robustly handle scenarios that are not very well represented in the training data.

\section{Conclusions \& Discussion}
\label{sec:conc}
We have provided two methods for training a neural network to predict its own correlated multivariate uncertainty: direct training with a Gaussian maximum likelihood loss function and indirect end-to-end training through a Kalman filter.
In addition, we have shown how to incorporate multivariate epistemic uncertainty during test time.
Our experiments show that these methods yield accurate uncertainty estimates and can dramatically improve the performance of a filter that uses them.
Significant improvement in filter state estimation came from accounting for both the heteroscedasticity in and correlation between the model outputs uncertainty.
For out-of-domain data, the incorporation of epistemic uncertainty was critical to the high performance of the combined filtering system.
Finally, when the data violated the underlying filter assumptions, the uncertainty estimates trained end-to-end through the filter were able to partially compensate for the resulting errors.
These methods of multivariate uncertainty estimation help enable the usage of neural networks in critical applications such as navigation, tracking, and pose estimation.

Our methods have several limitations that should be addressed by future work.
First, we have assumed that both aleatoric and epistemic uncertainty for neural network predictions follow multivariate Gaussian distributions.
While this is a reasonable assumption for many applications, it is problematic in two scenarios:
(1) When error distributions are long-tailed, the likelihood of large errors may be extremely underestimated;
(2) When errors follow a multimodal distribution, which can occur in highly nonlinear systems, the precision of the estimation will be impaired.
The former scenario may be addressed by replacing the Gaussian distribution with a Laplace distribution, which produces more stable training gradients for large errors.
The latter scenario may be addressed by quantile regression extensions to our method.
While both of these scenarios violate Kalman filter assumptions, they can be handled with more advanced Bayes filters such as particle filters.

Finally, our methods assume that the uncertainty for measurements is \emph{uncorrelated} in time between different measurements.
In most real-world applications, we expect uncertainty to be made up of both independent sources, such as sensor noise, and correlated sources, such as occluding scene content visible in multiple frames.
The independence assumption was strongly violated in our visual odometry experiment in Section \ref{sec:odometry}, which required a special Kalman filter variant to handle it.
A valuable extension to our work would allow neural networks to quantify and predict these correlations in a way that is interpretable and easy to integrate into conventional filtering systems.

\bibliographystyle{ieeetr}
\bibliography{references} 

\end{document}